\documentclass{article}
\usepackage{spconf,amsmath,graphicx}
\usepackage{algorithm}
\usepackage{algpseudocode}
\usepackage{tabularx}
\usepackage[hidelinks]{hyperref}
\usepackage{caption}

\title{Mask Consistency Regularization in Object Removal}%

\name{Hua Yuan$^{1,3}$, Jin Yuan$^{2}$, Yicheng Jiang$^{1}$, Yao Zhang$^{2}$, Xin Geng$^{1,3,\dagger}$, Yong Rui$^{1,\dagger}$\thanks{$^{\dagger}$Corresponding authors.}}
\address{
    $^{1}$School of Computer Science and Engineering, Southeast University, Nanjing, China \\
    $^{2}$AI Lab, Lenovo Research, Beijing, China \\
    $^{3}$Key Laboratory of New Generation Artificial Intelligence Technology and Its \\
    Interdisciplinary Applications (Southeast University), Ministry of Education, China \\
    \{yuanhua, ycjiang, xgeng\}@seu.edu.cn, \{jinyuan, yaozhang\}@lenovo.com, yongrui@outlook.com 
}

\begin{document}
\ninept
\maketitle

\begin{abstract}
Object removal, a challenging task within image inpainting, involves seamlessly filling the removed region with content that matches the surrounding context. Despite advancements in diffusion models, current methods still face two critical challenges: (1) mask hallucination, where the model generates irrelevant or spurious content inside the masked region, and (2) mask-shape bias, where the model fills the masked area with an object that mimics the mask's shape rather than surrounding content. To address these issues, we propose Mask Consistency Regularization (MCR), a novel training strategy designed specifically for object removal tasks. During training, our approach introduces two mask perturbations: dilation and reshape, enforcing consistency between the outputs of these perturbed branches and the original mask. The dilated masks help align the model’s output with the surrounding content, while reshaped masks encourage the model to break the mask-shape bias. This combination of strategies enables MCR to produce more robust and contextually coherent inpainting results. Our experiments demonstrate that MCR significantly reduces hallucinations and mask-shape bias, leading to improved performance in object removal.
\end{abstract}


\section{Introduction}
Image inpainting, particularly for object removal, aims to reconstruct missing regions of an image with visually realistic and semantically coherent content~\cite{pathak2016context, wu2025kris}. Early deep learning methods adopted convolutional encoder–decoder frameworks and generative adversarial networks (GANs) that leveraged attention mechanisms or partial convolutions to borrow textures from the surrounding context~\cite{yu2018generative, liu2018image}. Recent diffusion models, such as Stable Diffusion, have brought significant improvements in visual fidelity and diversity for inpainting~\cite{lugmayr2022repaint, rombach2022high}. Consequently, recent research has predominantly focused on diffusion-based approaches in pursuit of superior image restoration quality~\cite{mo2024dynamic, zhang2025decoupling}. 

In the context of object removal, existing approaches can be broadly categorized into prompt-based and mask-based methods. Prompt-based methods leverage text guidance to specify undesired objects, offering greater flexibility and usability in interactive editing scenarios. For example, Imagen Editor~\cite{chefer2023imageneditor} and InstructPix2Pix~\cite{brooks2023instructpix2pix} demonstrate that diffusion models can be steered by natural language instructions to remove or modify objects in a controllable manner. On the other hand, mask-based methods exploit explicit spatial priors, often in conjunction with segmentation models, to ensure more precise and reliable object removal. Notably, approaches such as LaMa~\cite{suvorov2022lama} and SmartBrush~\cite{liu2023smartbrush} integrate corrupted image regions and binary masks into the generative process, yielding structurally coherent and semantically faithful completions. These two lines of research highlight the trade-off between flexibility and accuracy in diffusion-based object removal.

However, despite these advances, two key challenges remain. First, \emph{mask hallucination} occurs when the model generates irrelevant or spurious objects within the masked region, introducing unnecessary content beyond the intended completion. Second, \emph{mask-shape bias} arises when the model interprets the geometric outline of the mask as a semantic cue, producing content that matches the silhouette rather than the surrounding scene. Even worse, the model may regenerate the removed object according to the mask shape, as illustrated in Fig.~\ref{fig:examples} in Section \ref{section:Experiments}. These issues highlight the key obstacles to effective object removal.

Inspired by the consistency regularization in the semi-supervised learning~\cite{sohn2020fixmatch}, we propose a simple yet effective training framework called \textbf{Mask Consistency Regularization} for diffusion-based object removal. The central idea is to enforce prediction consistency under mask perturbations. Specifically, during training, we introduce two perturbation strategies: (1) \emph{mask dilation}, which applies morphological dilation~\cite{soille1999morphological} to the original mask, enlarging its region to encourage alignment with the surrounding content, and (2) \emph{mask reshape}, which constructs random masks or rectangles that cover the original mask area, thereby reducing the model's dependency on the geometry of the mask. By enforcing consistency between outputs under perturbed and original masks, MCR enhances coherence around mask boundaries while regularizing the inpainting process against mask-specific artifacts.


\begin{figure*}[t]
  \centering
  \includegraphics[width=1.0\textwidth]{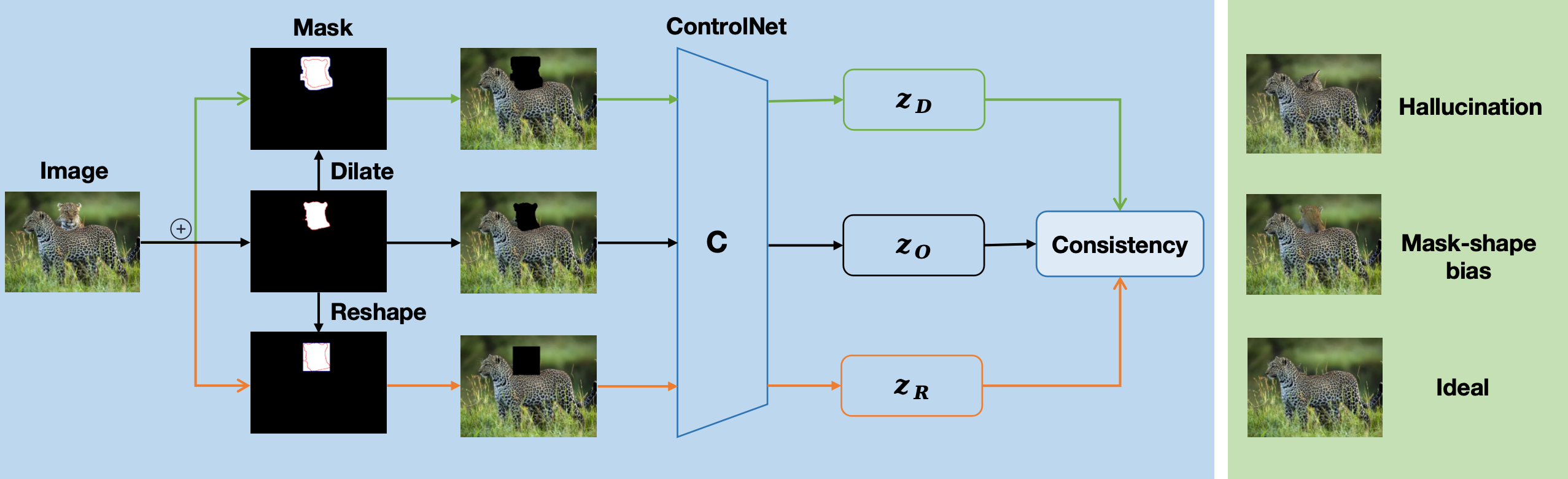}
  \caption{(1) The left part is the overall pipeline of our approach. Given an input image and its corresponding binary mask (outlined in red), we generate two variants of the mask through perturbation: mask dilation and mask reshaping. These perturbed masks, along with the original image, are processed by ControlNet to produce conditional latent encodings ($z_0, z_1, z_2$). A consistency regularization is enforced between the outputs generated from the original and perturbed masks, guiding the diffusion model to produce coherent and contextually consistent inpainting results. (2) The right part is the illustration of hallucination, mask-shape bias and an ideal output.}
  \label{fig:teaser}
\end{figure*}

The proposed method builds upon the Stable Diffusion XL model (SDXL)~\cite{podell2023sdxl} and employs a plug-and-play ControlNet module~\cite{zhang2023adding} to guide object removal, where the parameters of SDXL remain entirely frozen to substantially reduce memory usage and training time. 
Fig.~\ref{fig:teaser} illustrates the training framework of our proposed Mask Consistency Regularization, where perturbed masks (via dilation and reshaping) are applied to the input image. Each of these masked images is processed through ControlNet to produce conditional representations, and consistency regularization is enforced across their outputs, encouraging robust and geometry-agnostic inpainting.


Our contributions can be summarized as follows.
\begin{itemize}
    \item We identify two critical challenges in diffusion-based object removal, namely \emph{mask hallucination} and \emph{mask-shape bias} and propose an effective framework called Mask Consistency Regularization, which enforces prediction consistency under mask perturbations to alleviate these problems.
    \item We introduce two perturbation strategies: (1) \emph{mask dilation}, which applies morphological dilation to expand the original region and improve contextual alignment, and (2) \emph{mask reshape}, which constructs reshaped masks covering the original mask to reduce shape-dependent fillings.
    \item Extensive experiments demonstrate that MCR significantly reduces hallucinations and mask-shape bias compared with existing baselines, and further validate the effectiveness of the proposed two perturbation strategies. 
\end{itemize}

\section{Method}
\label{sec:method}
Our approach frames diffusion-based object removal as a consistency-driven learning problem. 
Instead of relying solely on reconstruction objectives, we systematically introduce perturbations to the input masks and require the model to produce stable and coherent predictions across these variations, as illustrated in Fig.~\ref{fig:teaser}. 
This design explicitly decouples the generation process from spurious dependencies on mask geometry, guiding the model to learn semantically meaningful completions rather than memorizing mask boundaries. 
By incorporating Mask Consistency Regularization into the inpainting framework, our method improves generalization, mitigates mask-induced artifacts, and enhances the overall quality of object removal.

\subsection{Preliminary}
We begin by clarifying the notations used in the task. 
Let \( x_0 \) denote the original image, \( M \) denote the binary mask, and \( \epsilon \) represent Gaussian noise added during the forward diffusion process. At a given timestep \( t \), the noisy sample is \( x_t \), obtained by perturbing \( x_0 \) with \( \epsilon \). 
A plug-and-play ControlNet module is employed to guide mask-based object removal. Specifically, the original image \( x_0 \) and the mask \( M \) are input to ControlNet, which produces a latent encoding \( z(x_0, M) \). This encoding, along with the text prompt \( p \), is then used as conditioning information to guide the SDXL model in generating the final image.
With these definitions, the reconstruction loss is given by:
\begin{align}
L_{\text{rec}} = \left\| \epsilon - \epsilon_\theta(x_0, t, z(x_0, M), p)  \right\|^2 .
\end{align}
This standard denoising objective drives the network to match the ground-truth noise under the given conditioning.

\subsection{Mask Perturbation Strategies}
Mask perturbations are central to our method, as they explicitly challenge the model to remain invariant to variations in mask geometry and coverage. Two perturbation strategies, mask dilation and mask reshape, are designed to reduce hallucinations and mask-shape bias.

\noindent\textbf{Mask Dilation.} 
Mask dilation is inspired by a classical morphological operation. 
Formally, let $M \in \{0,1\}^{H \times W}$ denote a binary mask, and let $K \in \{0,1\}^{(2k+1)\times(2k+1)}$ be a square structuring element filled with ones. 
The $k$-dilated mask, denoted as $M^{(k)}_{\text{dil}}$, is obtained via the morphological dilation operator $\otimes$:
\begin{align}
M^{(k)}_{\text{dil}} = M \otimes K,
\end{align}
where $\otimes$ represents the classical morphological dilation~\cite{gonzalez2009digital}. 
Intuitively, this operation enlarges the original masked region by $k$ pixels in all directions, thereby expanding its boundary and encouraging the model to account for contextual information beyond the strict mask edges.

\noindent\textbf{Mask Reshape.} 
In addition to dilation, we introduce \emph{mask reshape} to reduce the model’s reliance on mask geometry. Two instantiations are considered: rectangular reshape and random reshape. Rectangular reshape replaces the original mask with its minimum enclosing rectangle. Formally, let $M \in \{0,1\}^{H \times W}$ denote a binary mask. We define
\begin{align*}
i_{\min} &= \min \{\, i \mid \exists j, \; M_{i,j} = 1 \,\}, \\
i_{\max} &= \max \{\, i \mid \exists j, \; M_{i,j} = 1 \,\}, \\
j_{\min} &= \min \{\, j \mid \exists i, \; M_{i,j} = 1 \,\}, \\
j_{\max} &= \max \{\, j \mid \exists i, \; M_{i,j} = 1 \,\}.
\end{align*}
The reshaped rectangular mask $M_{\text{rect}}$ is then defined as
\begin{align}
[M_{\text{rect}}]_{i,j} = 
\begin{cases}
1, & \text{if } i_{\min} \leq i \leq i_{\max} \text{ and } j_{\min} \leq j \leq j_{\max}, \\
0, & \text{otherwise}.
\end{cases}
\end{align}
This operation produces the tightest rectangle aligned with the axis that fully covers the original mask region. Random reshape augments the original mask with an additional random mask. Following the random mask generation strategy proposed in LaMa~\cite{suvorov2022lama}, we first generate a random binary mask $M_{0} \in \{0,1\}^{H \times W}$. 
The reshaped mask $M_{\text{rand}}$ is then obtained as:
\begin{align}
M_{\text{rand}} = M \lor M_{0},
\end{align}
where $\lor$ denotes the element-wise OR operator. 
This operation perturbs the original region in a stochastic manner, ensuring the model cannot rely solely on the geometric shape of the provided mask. In our experiments, rectangular reshape and random reshape are applied with equal probability of 50\%.


\subsection{Mask Consistency Regularization}
Based on the two proposed mask perturbation strategies, we introduce the {Mask Consistency Regularization} algorithm. 
The corresponding latents generated by the ControlNet for these perturbed masks are \( z(x_0, M_{\text{dil}}^{(k)}) \) and \( z(x_0, M_{\text{reshape}}) \), which are abbreviated as \( z_D \) and \( z_R \), respectively. The latents generated by the original mask are abbreviated as \( z_O \).
To enforce mask consistency, we introduce the consistency loss function, which encourages the model to produce similar predictions for the original and perturbed masks. Specifically, the consistency loss is defined as:
\begin{align}
\mathcal{L}_{\text{cons}} &= \left\| \epsilon_\theta(x_0, t, z_O, p) - \epsilon_\theta(x_0, t, z_D, p) \right\|^2 \nonumber \\
&\quad + \left\| \epsilon_\theta(x_0, t, z_O, p) - \epsilon_\theta(x_0, t, z_R, p) \right\|^2.
\end{align}
This loss function forces the model to maintain consistency in its predictions when the mask is perturbed. By encouraging the model to produce similar latent representations for the original and perturbed masks, the model becomes less sensitive to variations in mask geometry and more focused on semantically meaningful content generation.
The overall training objective integrates the reconstruction loss with the  consistency regularization
\begin{align}
\label{eq:total_loss}
\mathcal{L} = \mathcal{L}_{\text{rec}} + \lambda_{\text{cons}} \, \mathcal{L}_{\text{cons}},
\end{align}
where \( \lambda_{\text{cons}} \) is a hyperparameter that balances the contribution of the consistency term. This design encourages the model not only to faithfully reconstruct realistic inpainted regions, but also to remain robust against variations in mask geometry, thereby enhancing generalization and improving the quality of object removal.

\begin{figure*}[t]
  \centering
  \includegraphics[width=\linewidth]{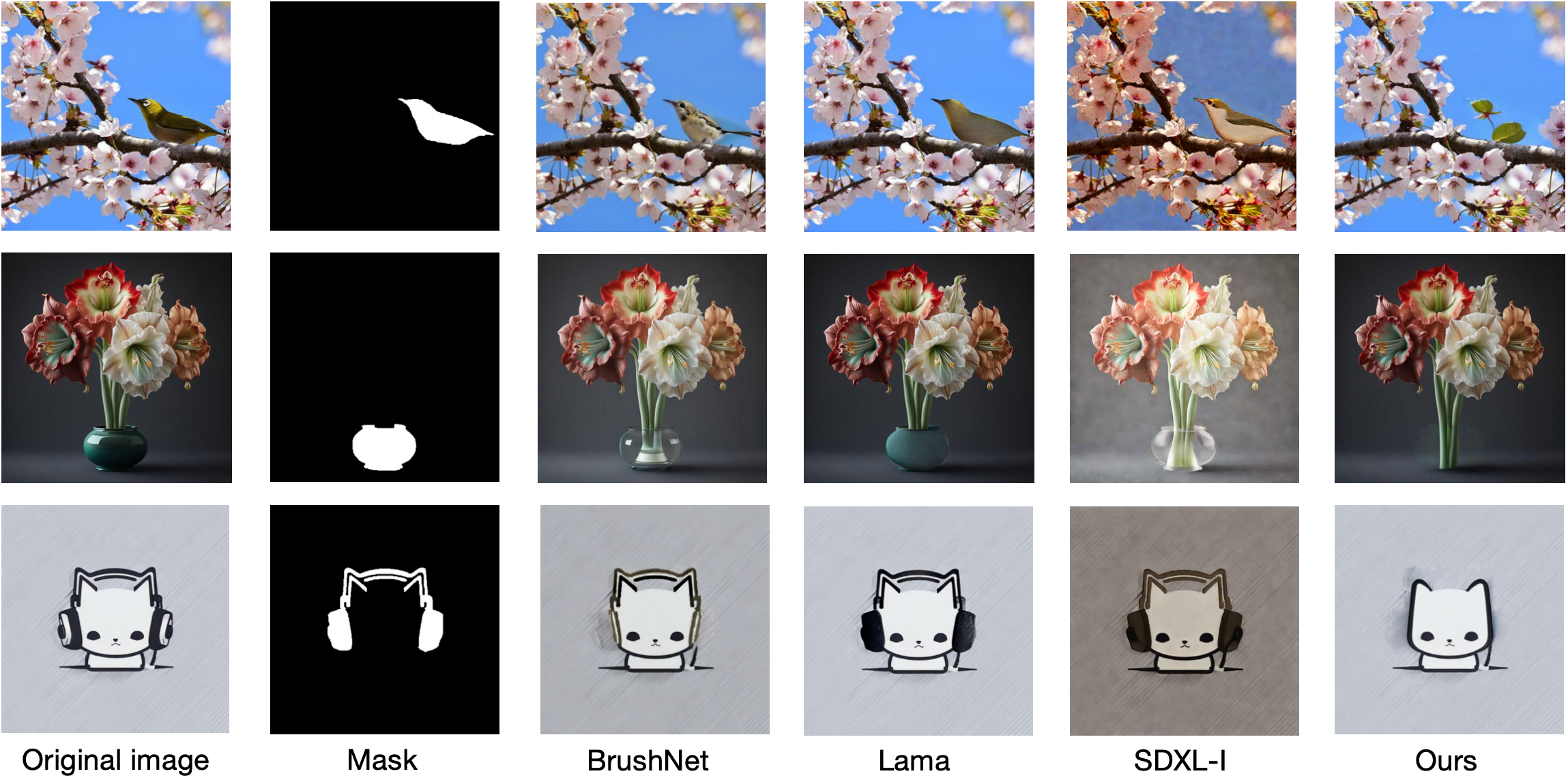}
  \vspace{0.1mm}
  \caption{Comparison of previous inpainting methods and ours on three examples from the BrushBench. In object removal, prior methods often hallucinate content or reconstruct the original object following the mask shape, while our approach effectively alleviates both issues.}

  \label{fig:examples}
\end{figure*}

\section{Experiments}
\label{section:Experiments}
We are in the process of conducting extensive experiments to evaluate the proposed method. In this section, we outline our experimental setup and show experimental results. We compared our method with previous approaches on multiple datasets using several metrics, and these results reflect the effectiveness of the proposed method. Additionally, we also present inpainting examples from each method, validating the improvements brought by MCR in mitigating model hallucinations and reducing shape-dependent fillings.

\subsection{Experimental Setup}
\noindent\textbf{Implementation Details.}
We adopt the Stable Diffusion XL model (SDXL) as the base model and employ a plug-and-play ControlNet module to guide object removal. During training, the parameters of SDXL are frozen, and only the parameters of ControlNet are updated. All training images are resized to a width of 1024 pixels. The training is conducted with a learning rate of $5 \times 10^{-5}$, a batch size of 2, and the consistency weight $\lambda_{\text{cons}}$ in Eq.~\ref{eq:total_loss} is set to 2. The training is performed on 8 NVIDIA H20 GPUs.

\noindent\textbf{Benchmark.} 
We evaluate our Mask Consistency Regularization on two benchmarks. The first is BrushBench~\cite{ju2024brushnet}, a segmentation-based inpainting dataset with 600 images annotated with human-annotated masks and captions. The second is the 300-sample removal test, a self-constructed dataset comprising real-world images, physics-engine generated scenes, and hand-drawn illustrations. Unlike prior benchmarks, it provides ground-truth images after removal, either human-annotated or synthesized by the physics engine, enabling more reliable evaluation. Since existing inpainting datasets generally lack such ground-truths, constructing dedicated datasets is a common practice~\cite{yu2025omnipaint, wei2025omnieraser}.

\noindent\textbf{Metrics.} We evaluate all methods with five metrics: FID~\cite{heusel2017gans}, PSNR~\cite{wang2004image}, SSIM~\cite{wang2004image}, LPIPS~\cite{zhang2018unreasonable} and CMMD~\cite{jayasumana2024rethinking}. FID {(↓)} measures distribution-level fidelity. PSNR {(↑)} and SSIM {(↑)} assess pixel-wise and structural reconstruction accuracy. LPIPS {(↓)} captures learned perceptual similarity and CMMD {(↓)} evaluates CLIP-based distribution discrepancy. This set of metrics jointly reflects statistical fidelity, perceptual quality, and mask-invariance behavior under our consistency regularization.

\noindent\textbf{Baselines.} We compare our method against several baselines: {LaMa}~\cite{suvorov2022lama} {BrushNet}~\cite{ju2024brushnet} (dual-branch diffusion inpainting), and {SDXL-I}~\cite{sdxl_inpainting}. For each baseline, we adopted the pre-trained model parameters and inference code released by them. In addition, the inference step size of all diffusion models was set to 20. For fairness, all methods are tested under the same conditions on our evaluation masks.

\begin{table}[h]
\centering
\captionsetup{skip=3pt} 
\caption{Evaluation on BrushBench.}
\begin{tabularx}{\linewidth}{lccccc}
\hline
Model & FID$\downarrow$ & PSNR$\uparrow$ & SSIM$\uparrow$ & LPIPS$\downarrow$ & CMMD$\downarrow$ \\
\hline
\textbf{BrushNet} & 63.55 & 17.60 & 0.7424 & 0.2063 & 0.1847 \\
\textbf{LaMa} & \textbf{58.95} & 22.08 & 0.8661 & 0.1258 & 0.1136 \\
\textbf{SDXL-I} & 120.74 & 15.29 & 0.5584 & 0.6587 & 0.8916 \\
\textbf{Ours} & 60.89 & \textbf{23.54} & \textbf{0.8969} & \textbf{0.1218} & \textbf{0.0741} \\
\hline
\end{tabularx}
\label{table:1}
\end{table}

\begin{table}[h]
\centering
\captionsetup{skip=3pt} 
\caption{Evaluation on the 300-sample removal test.}
\begin{tabularx}{\linewidth}{lccccc}
\hline
Model & FID$\downarrow$ & PSNR$\uparrow$ & SSIM$\uparrow$ & LPIPS$\downarrow$ & CMMD$\downarrow$ \\
\hline
\textbf{BrushNet} & 61.66 & 21.33 & 0.7537 & 0.1492 & 0.0602 \\
\textbf{LaMa} & 33.98 & 29.09 & \textbf{0.8844} & 0.0916 & 0.0525 \\
\textbf{SDXL-I} & 37.41 & 28.53 & 0.8675 & 0.1040 & 0.0618 \\
\textbf{Ours} & \textbf{30.35} & \textbf{29.69} & 0.8666 & \textbf{0.0835} & \textbf{0.0428} \\
\hline
\end{tabularx}
\label{table:2}
\end{table}

\subsection{Experimental Results}
We evaluate the proposed method using multiple performance metrics across two distinct benchmarks. The results of these evaluations are shown in Table~\ref{table:1} and Table~\ref{table:2}. As indicated by the tables, our method outperforms the baselines on most metrics. It demonstrates its ability to preserve both visual fidelity and structural accuracy. Additionally, our method achieves the best LPIPS and CMMD scores, which further highlight our method's superior perceptual quality and robustness against mask-induced artifacts. These results underscore the effectiveness of Mask Consistency Regularization in reducing hallucinations and improving object removal performance.

\begin{table}[h]
\centering
\captionsetup{skip=3pt}
\caption{Ablation studies on BrushBench.}
\begin{tabularx}{\linewidth}{lccccc}
\hline
Model & FID$\downarrow$ & PSNR$\uparrow$ & SSIM$\uparrow$ & LPIPS$\downarrow$ & CMMD$\downarrow$ \\
\hline
MCR & \textbf{60.89} & \textbf{23.54} & \textbf{0.8969} & \textbf{0.1218} & \textbf{0.0741} \\
w/ $M_D$   & 63.83 & 21.81 & 0.8683 & 0.1539 & 0.0994 \\
w/ $M_R$   & 62.33 & 21.65 & 0.8681 & 0.1531 & 0.0832 \\
\hline
\end{tabularx}
\label{table:3}
\end{table}

\begin{table}[h]
\centering
\captionsetup{skip=3pt} 
\caption{Ablation studies on the 300-sample removal test.}
\begin{tabularx}{\linewidth}{lccccc}
\hline
Model & FID$\downarrow$ & PSNR$\uparrow$ & SSIM$\uparrow$ & LPIPS$\downarrow$ & CMMD$\downarrow$ \\
\hline
MCR & \textbf{30.348} & \textbf{29.692} & \textbf{0.8666} & \textbf{0.0835} & \textbf{0.0428} \\
w/ $M_D$ & 33.684 & 29.227 & 0.8654 & 0.0872 & 0.0437 \\
w/ $M_R$ & 32.016 & 29.472 & 0.8659 & 0.0862 & 0.0441 \\
\hline
\end{tabularx}
\label{table:4}
\end{table}

Figure~\ref{fig:examples} provides a visual comparison of inpainting results between our method and several baselines. It shows that previous inpainting methods are prone to generating hallucinated content. As illustrated in the figure, these baselines often generate hallucinations based on the shape of the mask and produce new objects within the masked region. In contrast, our method significantly reduces hallucinations and shape-dependent fillings. The results demonstrate that our MCR framework, by enforcing consistency across perturbed masks, enables the diffusion model to generate more coherent and contextually appropriate inpainting results. Our method does not rely on mask shape, which further contributes to more reliable object removal and reduces artifacts typically observed in mask-based approaches. These results emphasize the advantages of our approach in both mitigating hallucinations and improving mask-agnostic inpainting.


\subsection{Ablation Studies}
We perform ablation studies on two mask perturbations across two benchmarks. The results on BrushBench are shown in Table~\ref{table:3}, while the results on our self-constructed 300-sample removal test set are shown in Table~\ref{table:4}. We use w/ $M_D$ to denote the consistency regularization using only mask dilation and w/ $M_R$ to denote the consistency regularization using only mask reshape. The results demonstrate that using both perturbations for object removal yields the best performance, which also indicates that both of the proposed mask perturbations are effective.

\section{Conclusion}
In this paper, we proposed the Mask Consistency Regularization framework to address the challenges of mask hallucination and mask-shape bias in object removal. By introducing mask dilation and mask reshaping strategies, MCR encourages the model to generate more contextually coherent and shape-agnostic inpainting results. Experimental results also demonstrate that our approach significantly improves visual fidelity, structural accuracy, and perceptual quality compared to existing baselines. Furthermore, MCR effectively reduces artifacts related to mask geometry, making the inpainting process more robust. In conclusion, the proposed Mask Consistency Regularization offers a promising direction for enhancing diffusion-based inpainting models, particularly in tackling the challenges related to mask-induced artifacts. Future work will further explore the impact of different perturbation strategies and find more effective strategy.

\bibliographystyle{IEEEbib}
\bibliography{Template}

\end{document}